\documentclass[sigconf]{acmart}
\AtBeginDocument{%
  \providecommand\BibTeX{{%
    \normalfont B\kern-0.5em{\scshape i\kern-0.25em b}\kern-0.8em\TeX}}}

\setcopyright{acmcopyright}
\copyrightyear{2018}
\acmYear{2018}
\acmDOI{XXXXXXX.XXXXXXX}

\usepackage{mathtools}
\usepackage{graphicx}
\usepackage{wrapfig}
\usepackage{multirow}
\usepackage{subcaption}
\usepackage{algorithm}
\usepackage{algorithmic}
\usepackage{booktabs}
\usepackage{balance}

\begin{document}

\title{Explainable and Accurate Natural Language Understanding for Voice Assistants and Beyond}


\author{Kalpa Gunaratna}
\affiliation{%
  \institution{Samsung Research America}
  \city{Mountain View, CA}
  \country{USA}}
\email{k.gunaratna@samsung.com}

\author{Vijay Srinivasan}
\affiliation{%
  \institution{Samsung Research America}
  \city{Mountain View, CA}
  \country{USA}}
\email{v.srinivasan@samsung.com}

\author{Hongxia Jin}
\affiliation{%
  \institution{Samsung Research America}
  \city{Mountain View, CA}
  \country{USA}}
\email{hongxia.jin@samsung.com}

\renewcommand{\shortauthors}{Kalpa Gunaratna, Vijay Srinivasan, \& Hongxia Jin}


\begin{abstract}
Joint intent detection and slot filling, which is also termed as joint NLU (Natural Language Understanding) is invaluable for smart voice assistants. Recent advancements in this area have been heavily focusing on improving accuracy using various techniques. Explainability is undoubtedly an important aspect for deep learning-based models including joint NLU models. Without explainability, their decisions are opaque to the outside world and hence, have tendency to lack user trust. Therefore to bridge this gap, we transform the full joint NLU model to be `inherently' explainable at granular levels without compromising on accuracy. Further, as we enable the full joint NLU model explainable, we show that our extension can be successfully used in other general classification tasks. We demonstrate this using sentiment analysis and named entity recognition. 

\end{abstract}




\maketitle

\section{Introduction}
Natural Language Understanding (NLU) is a critical component in building intelligent interactive agents such as Amazon Alexa, Google Assistant, Samsung's Bixby, and Microsoft's Cortana. In order to complete user requests, these virtual assistants need to understand intents and slots. Intents reflect the actions need to perform and slots represent the entity phrases that are required to complete the actions. For example, for the utterance "Book me a flight from Denver to New Jersey", intent can be "book\_flight" and slots (class-value pairs) can be (origin-Denver) and (destination-New Jersey). Intent detection (i.e., intent classification) is handled by classifying the input utterance into one of the pre-determined intent classes and slot filling (i.e., slot classification) is about classifying each and every token (or a sequence of tokens) of the input utterance into one of the pre-determined slot classes. Slot filling is typically handled through sequence labeling-based (BIO notation) classification (e.g., ~\cite{gunaratna2021using}).

Jointly optimizing the above two tasks has been shown to be the optimal approach~\cite{zhang2019joint,qin2021co}. Some existing approaches perform some type of feature learning but they all do it at coarser level (e.g., ~\cite{zhang2019joint, qin2021co, he2020syntactic}) than granular, class-specific level. These systems make use of different encoding mechanisms such as RNN~\cite{liu2016attention,goo2018slot}, CNN~\cite{xu2013convolutional}, intent-based attention~\cite{dao2021intent}, stack propagation~\cite{qin2019stack}, and hierarchical information flow~\cite{lee2018coupled,zhang2019joint,zhang2020graph}. On the other hand, explainability has been a hot topic in natural language processing (NLP) and machine learning communities~\cite{danilevsky2020survey} as it can enable the model to be transparent of its decisions and gain user trust. Post-hoc processing techniques can be applied to most of the learning methods but they are considered alien to the model and hence, argued to be problematic and not trustworthy~\cite{rudin2019stop}. Therefore, lot of interest has been on the inherently explainable models where, explainability is integral and built into the model. In this work, we show that both improving accuracy through fine-grained feature learning and introducing granular level inherent explainability can be achieved for the full joint NLU model. Fine-grained feature computation for slots has been shown to be possible~\cite{gunaratna2022explainable}. However, no investigation has been done for intents and then for the full joint NLU model that has both intents and slots. Most importantly, we show that, our method can be used in other classification tasks to enable inherent model explanibility. That is, our approach is not confined to joint NLU but can be used in other classification tasks.

\section{Approach}

\begin{figure*}
    \centering
    \includegraphics[scale=0.65, trim=0cm 8cm 1.2cm 1cm, clip=true]{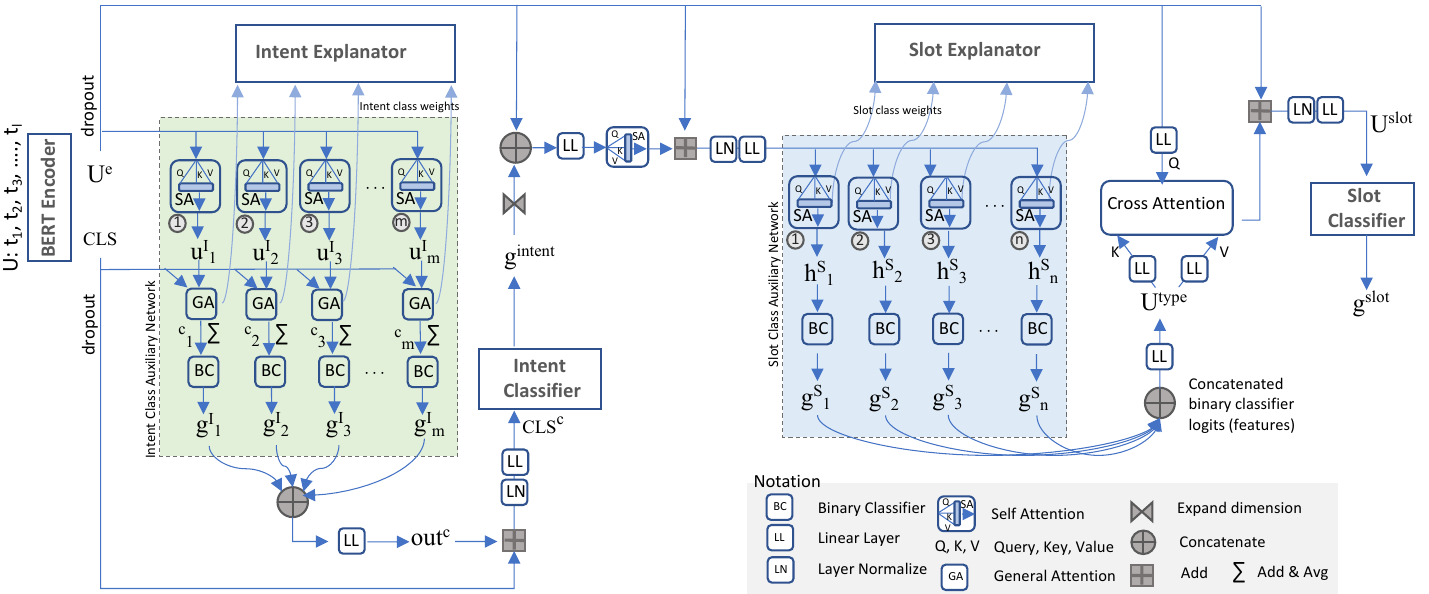}
    \caption{Overview of our explainable joint NLU model. n and m are the total number of slot and intent classes, l is the utterance length. In addition to intent and slot predictions, our model visualizes per slot and intent class attentions as explanations.}
    \label{fig_architecture}
    \vskip -0.1in
\end{figure*}

Our method for improving both classification accuracy and inherent explainability is based on fine-grained attention weights computation. We use properly computed attention weights to explain model decisions. Wiegreffe et al.~\cite{wiegreffe2019attention} showed that attentions can be used for model explanation purposes since attention mechanisms capture important details of the model decisions. Our model architecture for the joint NLU model is shown in Figure~\ref{fig_architecture}. Given the space limitation, we try to outline our approach with details below.

\paragraph{Preliminaries} Let $l$ be the number of tokens in the input utterance $u$=[$t_1$, $t_2$, .., $t_l$] where $t_i$ represents the $i$th token in $u$. Let $u^e$=$[\boldsymbol{t_1}, \boldsymbol{t_2}, ..., \boldsymbol{t_l}]$ ($\in \mathbb{R}^{l \times d}$) be the encoded utterance where $\boldsymbol{t_i}$ represents the embedding for the $i$th token, $\boldsymbol{t_i} \in \mathbb{R}^d$ and $d$ is the embedding dimension. Let $T$ be the slot class set in the ground truth (e.g., location, time, etc.), $S$ be the set of sequence (BIO format) labels (e.g., O, B-location, I-location, etc.), and $I$ be the set of intent classes in the ground truth (e.g., book\_flight, search, etc.). Given an utterance $u$, intent detection is to classify $u$ to one of intent classes from $I$ and slot filling is to classify single or consecutive set of tokens $t_1, .., t_x$ $\in u$ where $x \leq l$ to one of classes in $T$.

Our task is joint optimization of intent and slot classification. Moreover, one of our important goals is to be able to explain both the intent and slot classifiers. To enable this capability, we introduce two auxiliary networks. One for the intent classification (green color region in Figure~\ref{fig_architecture}) and the other is for slot classification (blue color region in Figure~\ref{fig_architecture}). Since we want to explain intent classification for each intent class for a given utterance, we want to learn to compute attention weights specific to each class. General query-based attention can be used to learn focus points in the utterance for intents. However, computing multiple query-based attentions (latter part of Equation~\ref{eq:intent_attn}) does not work as the utterance representation used is the same. Hence, we first transform the utterance into each class-specific representation using self-attention, one per intent class (1 to $m$ number of self-attentions shown in the Figure~\ref{fig_architecture}) as also mentioned in Equation~\ref{eq:intent_attn}. Then, on each class specific transformed utterance ([$u^{I}_1$, ..., $u^{I}_{m}$]), we compute query attentions ($GA$ blocks in Figure~\ref{fig_architecture}) to get the intent class-specific attention weights. These computed per-class attention weights $\alpha^I$ are fed into the intent explanator to explain intent classifications. Utterance transformation using self-attention, $score$ computation for each intent class using query-attention, and then attention weights $\alpha^{I}$ computation for each intent class is performed as shown below. $Q$, $K$, and $V$ are query, key, and value vector projections from the input utterance, $CLS$ is the utterance embedding, and $d$ is the dimension.

\begin{equation}
\begin{aligned}
    u_{i}^{I} = softmax \left(\frac{Q_{i} K_{i}^{T}}{\sqrt{d}} \right) V_{i} \\
    score_{i} = CLS \times u^{I}_{i}, \: \: \: \alpha_{i}^{I} = softmax(score_i)
\label{eq:intent_attn}
\end{aligned}
\end{equation}

However, these attention computations need to be further constrained to make them learn properly. Therefore, we perform binary classification ($BC$ blocks in Figure~\ref{fig_architecture}) over the averaged general attention weighted utterances  ([$c_1$, ..., $c_m$]) where $c_i$= $\sum_{x}^{l} \alpha_{i,x}^{I} u^I_{i,x}$ / $l$, one per class to get binary logits $g^I$ as shown in Equation~\ref{eq:logits}. $|W^I|$ = $|b^I|$ = $|I|$ and binary classifier logit $g^I_{i}$ $\in \mathbb{R}^1$ for $i \in I$.

\begin{equation}
    g^I_{i} = c_i W^I_{i} + b^I_{_i}
    \label{eq:logits}
\end{equation}

Output logits of these binary classifiers are used to compute binary cross entropy loss $\mathcal{L}^X$. We then concatenate these binary logits that also represent high level patterns related features for intents, linearly transform to get contextual representation $out^c$. Then we add the original utterance representation $CLS$ to $out^c$, perform layer normalization and linear transformation to get $CLS^c$ and feed it into the main network's intent classifier to get the final intent predictions through intent logits $g^{intent}$. Intent logits are computed similarly to Equation~\ref{eq:logits}, where $g^{intent} = CLS^c \:  W^{intent} + b^{intent}$, and $W^{intent} \in \mathbb{R}^{d \times |I|}$. Intent loss $\mathcal{L}^{intent}$ is computed using cross entropy loss.

Similarly, we can enable slot class-specific fine-grained explanations. Slot classification is performed on each token in the input utterance. Hence, we provide explanations for each token classification with respect to all candidate slot classes. Similar to intents, we first transform the input utterance into multiple representations ([$h^S_{1}$, ..., $h^S_{n}$]) using self-attentions, exactly $n$ times, which is the number of slot classes. Importantly, self-attention weights ($\alpha^S$) resulting from this transformation are used for slot explanations. Self attention weights $\alpha^{S}$ and slot class-specific representations $h^S$ are computed as shown in Equation~\ref{eq:type_transform} below. $Q_s$, $K_s$, and $V_s$ are linear projections (for slot $s \in T$) of input and $d_h$ is the dimension of projected vector $K_s$. Input to this component is by concatenating intent logits $g^{intent}$ with original utterance $u^e$, linear projection into a self attention and then adding a residual connection followed by layer normalization and linear transformation as shown in Figure~\ref{fig_architecture}.

\begin{equation}
\alpha_{s}^{S} = softmax \left( \frac{Q_{s} K_{s}^{T}}{\sqrt{d_h}} \right), \: h^{S}_{s} 
    = \alpha_{s}^{S} V_{s}
\label{eq:type_transform}
\end{equation}

We expect these transformations to represent each and every slot class. To ensure this exact transformation, we enforce it through $n$ binary classifiers. Each binary classifier gets the respective transformed utterance (e.g., transformation $h^S_{s}$ for sth slot class) and produces binary logits $g^S$ (e.g., $g^S_{s}$ for slot class $s$), reflecting each token belongs to that slot class or not, where $g^{S}_{s} = h^{S}_{s} W^S_{s} + b^S_{s}$ and $|W^S|$ = $|b^S|$ = $|T|$. $h^{S}_{s} \in \mathbb{R}^{l \times d_h}$, $W^S_{s} \in \mathbb{R}^{d_h \times 1}$, $b^S_{s} \in \mathbb{R}$, and slot class logits $g^{S}_{s} \in \mathbb{R}^{l \times 1}$. Binary classification loss $\mathcal{L}^Y$ is computed using binary cross entropy. Then these binary logits are concatenated, linearly transformed, and infused with the input utterance embeddings using cross-attention to extract slot class-specific features from the input utterance. Then finally, we add a residual connection, followed by layer normalization and linear projection to these extracted features to get $u^{slot}$ and feed into the slot classifier to predict slot logits ($g^{slot}$ $\in \mathbb{R}^{l \times |S|}$), where $g^{slot} = u^{slot} W^{slot} + b^{slot}$, from which the maximum probability prediction is selected as the slot for an input token. Weights vector is $W^{slot}$ ($\in \mathbb{R}^{d \times |S|}$). We use cross entropy to compute slot classifier loss $\mathcal{L}^{slot}$. Then our entire network optimization is performed using total loss $\mathcal{L}$ = $\lambda$ $\mathcal{L}^{intent}$ + $\beta$ $\mathcal{L}^X$ + $\gamma$ $\mathcal{L}^Y$ + $\eta$ $\mathcal{L}^{slot}$, where $\lambda, \beta, \gamma, \eta$ are loss weights.
Intent and slot explanations at granular levels are performed using class-specific attention weights computed in Equations~\ref{eq:intent_attn} and ~\ref{eq:type_transform}.

\section{Evaluation and Discussion}

\begin{figure*}[]
    \centering
    \includegraphics[scale=0.73, trim=0cm 9cm 3cm 0cm, clip=true]{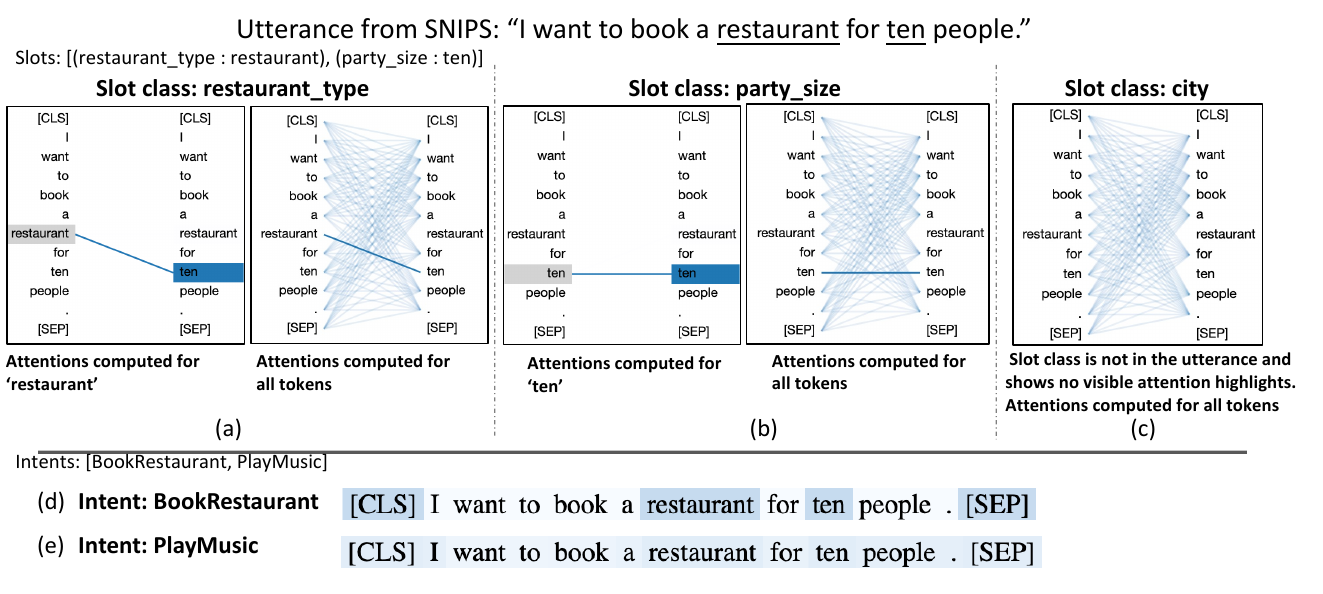}
    \caption{Slot and intent class-specific attentions and explainable visualizations on an example utterance from SNIPS. The top section illustrates slots and the bottom intents. (a) and (b) show plots for two `positive' slot classes (that appear in the utterance) whereas (c) shows a plot for a `negative' slot class. Blue rectangles in the left plots in (a) and (b) show the attention points of the model for the tokens in the gray rectangles for the target slot class. (d) and (e) show attention weights visualization for intents on the utterance. (d) shows the `positive' intent for the utterance whereas (e) represents a `negative' intent for the utterance.}
    \label{fig_visualize}
\end{figure*}

\begin{table}[]
\centering
\small
\begin{tabular}{lllll}
\hline
\multirow{2}{*}{Model}                          & \multicolumn{2}{c}{SNIPS}    & \multicolumn{2}{c}{ATIS}      \\ \cline{2-5} 
                                                & Intent & Slot & Intent & Slot \\ \hline
RNN-LSTM~\cite{hakkani2016multi}                & 96.9              & 87.3      &  92.6             & 94.3                  \\
Attention-BiRNN~\cite{liu2016attention}         & 96.7              & 87.8      &  91.1             & 94.2                  \\
Slot-Gated~\cite{goo2018slot}                   & 97.0              & 88.8      &  94.1             & 95.2                  \\ \hline
Joint BERT~\cite{chen2019bert}                  & 98.6              & 97.0      &  97.5             & 96.1                  \\
Joint BERT + CRF~\cite{chen2019bert}            & 98.4              & 96.7      &  97.9             & 96.0                  \\
Co-Inter. Transformer~\cite{qin2021co}        & 98.8              & 95.9      &  97.7             & 95.9                  \\
Co-Inter. Transformer + BERT~\cite{qin2021co} & 98.8              & 97.1      &  98.0             & 96.1                  \\ \hline
Slot explanation only~\cite{gunaratna2022explainable}                 & 98.99     & \textbf{97.24} &  99.10& \textbf{96.20} \\
Full joint NLU explainable model                 & \textbf{99.14}     & \textbf{97.24} &  \textbf{99.28}    & 96.19\\

\hline
\end{tabular}
\caption{Joint intent detection and slot filling results.
}
\label{tab:eval_accuracy}
\end{table}

We evaluate our joint NLU model against other baseline methods using SNIPS~\cite{coucke2018snips} and ATIS~\cite{hemphill1990atis} datasets.  We experimented with 20, 30, and 40 epochs, batch sizes of 16, 32, and 64, and trained the model for the entirety of the epochs.  We used learning rate of 5e-5, dropout of 0.1, and Adam optimizer. We used 0.5 and 1 for $\lambda$, $\beta$, $\gamma$, and $\eta$ and the slot class network attention (projection) dimension $d_h$=32. The results are shown in Table~\ref{tab:eval_accuracy}. It shows that our fully inherently explainable joint NLU model performs better than other comparable state-of-the-art non-explainable baselines. We can also see that by adding intent explainability and additional intent-specific features, the model accuracy remained the same for slots whereas for intents, it improved over slot features/explanation only model. Accuracy improvements over existing state-of-the-art methods like Joint BERT and co-interactive transformer methods are mainly due to fine-grained feature computation and fusion that also enable inherent explainability for the full joint NLU model.

To evaluate fine-grained inherent explainability, we analyze entropy to check attention spikes for positive classification classes over negative classification classes.
That is, when the model makes a classification decision, attention weights need to be able to reflect that decision. Hence, positive classes in a classification instance should have non-uniform attention weights compared to negative classes (classes that are not true for current prediction) for the same instance. See Figure~\ref{fig_visualize} for few example fine-grained class-specific attention visualizations for both the tasks using an utterance from SNIPS. For slots, Figure~\ref{fig_visualize} (a) and (b) show positive slot classes and their attention weights for the utterance whereas (c) shows attention weights for a negative slot class for the same utterance. We can see that positive slot classes have clear attention weight spikes that shows corresponding model focus areas for those slot classes. We also see very similar behavior for the intents as visualized in Figure~\ref{fig_visualize} (d) and (e) where, (d) is the positive intent class for the utterance and (e) is one of negative intent classes for the utterance.

\begin{table}[]
\setlength\tabcolsep{2.5pt}
\begin{tabular}{r|lll|lll}
\hline
\multicolumn{1}{c|}{\multirow{2}{*}{\begin{tabular}[c]{@{}c@{}}Top $k$\% \\ attn.\end{tabular}}} & \multicolumn{3}{c|}{SNIPS} & \multicolumn{3}{c}{ATIS} \\ \cline{2-7} 
\multicolumn{1}{c|}{}                                                                                & Pos.    & Neg.    & Diff   & Pos.   & Neg.   & Diff   \\ \hline
\multicolumn{7}{c}{Slot}      \\ \hline
100\%                                                                                          & \textbf{6.1161}  & 6.1933  & 0.0772 & \textbf{6.3135} & 6.4009 & 0.0874 \\
10\%                                                                                                 & \textbf{3.7597}  & 4.0000  & 0.2403 & \textbf{3.4835} & 3.7895 & 0.3060 \\
5\%                                                                                                  & \textbf{2.7458}  & 2.9941  & 0.2483 & \textbf{2.3836} & 2.7663 & 0.3827 \\ \hline
\multicolumn{7}{c}{Intent}      \\ \hline
100\%   & \textbf{2.9092}  & 3.5431  & 0.6339 & \textbf{2.4051} & 3.2529 & 0.8478 \\ \hline
\end{tabular}
\caption{Average entropy for top $k$\% attention weights for test data. Lower entropy means non-uniform values. Results confirm attention spikes in Pos compared to Neg. Positive (Pos.) are the slot classes that appear in an utterance.
}
\label{tab:entropy}
\end{table}

We compute entropy as shown in Equation~\ref{eq:entropy} to evaluate all the test examples for the above behavior. Entropy reflects smaller values for non-uniform collection of values and hence, the positive classes in our examples should have lower entropy, if the method learns attention weights successfully. Table~\ref{tab:entropy} shows entropy values computed for both intent detection and slot filling tasks in the joint NLU model using SNIPS and ATIS datasets. We see that positive classes have lower average entropy values confirming our hypothesis. Note that, we also investigate top $k\%$ attention weights for slots to distinguish the difference between the positive and negative classes because, there are large number of attention values and hence, few attention spikes in a very long list of values may not reflect the difference. Entropy is computed as below.

\begin{equation}
    entropy = - \sum Px_i \times log_2(Px_i)
\label{eq:entropy}
\end{equation}
where, for a list of attention weights [$x_1$, $x_2$,.., $x_n$], $Px_i$ = $x_i$/$\sum x_i$

\subsection*{General Applicability for Inherent Explainability}
Our inherently explainable joint NLU model successfully provides fine-grained intent and slot explanations, without needing to consult any post-hoc processing techniques. We see that, intent detection and slot filling classification tasks represent many other general classification problems. Intent detection is about classifying the entire input into one of pre-determined classes and hence resembles problems like sentiment analysis and sentence classification. On the other hand, slot filling is about classifying each and every token/word in the input similar to named entity recognition (NER) and part of speech tagging. Therefore, our inherently explainable modules for intent detection and slot filling can be applied to other general tasks. To evaluate such general applicability, we select two representative tasks: sentiment analysis and NER.

We use two major benchmarks for these tasks: SST2~\cite{sst2} for sentiment analysis and conll2003~\cite{conll2003} for NER. To show that, we integrate our explainability inducing auxiliary networks into a BERT-based classification baseline method. The results are shown in Table~\ref{tab:additional_tasks}. It shows that, incorporating our explainability extensions does not introduce any model performance reduction. Note that, Du et al.~\cite{du2019techniques} mentioned that model accuracy drops in most cases when explainability is incorporated. We further evaluated the explainability functinality using entropy and the results are shown in Table~\ref{tab:entropy_general_tasks}. We can see from the results that similar to joint NLU task, positive classes in both the tasks have lower entropy compared to negative classes. We see that for sentiment analysis, the difference is a bit low, because there are only two classes in the sentiment analysis. This evaluation suggests that our explainability components can be successfully integrated into other classification problems to enable inherent explainability.

\begin{table}[]
\small
\centering
\begin{tabular}{lll}
\hline
Task                                & BERT          & BERT + Ours \\ \hline
NER (F1 Score) & 90.56         & \textbf{90.94}       \\
Sentiment Analysis (Accuracy)       & 92.09         & \textbf{92.14}       \\ \hline
\end{tabular}
\caption{Adding our fine-grained class-specific feature learning does not compromise the model performance.}
\label{tab:additional_tasks}
\end{table}

\begin{table}[]
\centering
\begin{tabular}{llll}
\hline
Top k \% & Positive & Negative & Difference \\ \hline
\multicolumn{4}{c}{NER}      \\ \hline
100\%    & \textbf{7.1190}     & 7.2897     & 0.1707      \\
10\%     & \textbf{4.2275}     & 4.4097     & 0.1822      \\
5\%      & \textbf{3.1709}     & 3.4063     & 0.2354      \\ \hline
\multicolumn{4}{c}{Sentiment}    \\ \hline
100\%    & \textbf{4.4229}     & 4.4278     & 0.0049      \\ \hline
\end{tabular}
\caption{Entropy scores for NER and sentiment analysis using conll2003 and SST2 datasets, respectively. Positive category entropy is lower than negative category in both tasks meaning, the model captures attention spikes for model decisions.}
\label{tab:entropy_general_tasks}
\end{table}

\section{Conclusion}
In this work, we showed that both intent detection and slot filling tasks in the joint NLU problem can be made inherently explainable without compromizing the accuracy. This makes the full joint NLU model explainable. Further, we showed that our method to enable explainability in intent and slot classifications can be successfully extended to other general classification tasks using two representative tasks: sentiment analysis and NER. This is an important finding as enabling inherent explainability in other classification tasks in general is invaluable. Further, task explanations can provide insights to model developers to debug and collect only the required data cost-effectively to improve the models.

\bibliographystyle{ACM-Reference-Format}
\balance
\bibliography{cikm-nlu-bibliography}


\begin{thebibliography}{22}


\ifx \showCODEN    \undefined \def \showCODEN     #1{\unskip}     \fi
\ifx \showDOI      \undefined \def \showDOI       #1{#1}\fi
\ifx \showISBNx    \undefined \def \showISBNx     #1{\unskip}     \fi
\ifx \showISBNxiii \undefined \def \showISBNxiii  #1{\unskip}     \fi
\ifx \showISSN     \undefined \def \showISSN      #1{\unskip}     \fi
\ifx \showLCCN     \undefined \def \showLCCN      #1{\unskip}     \fi
\ifx \shownote     \undefined \def \shownote      #1{#1}          \fi
\ifx \showarticletitle \undefined \def \showarticletitle #1{#1}   \fi
\ifx \showURL      \undefined \def \showURL       {\relax}        \fi
\providecommand\bibfield[2]{#2}
\providecommand\bibinfo[2]{#2}
\providecommand\natexlab[1]{#1}
\providecommand\showeprint[2][]{arXiv:#2}

\bibitem[Chen et~al\mbox{.}(2019)]%
        {chen2019bert}
\bibfield{author}{\bibinfo{person}{Qian Chen}, \bibinfo{person}{Zhu Zhuo},
  {and} \bibinfo{person}{Wen Wang}.} \bibinfo{year}{2019}\natexlab{}.
\newblock \showarticletitle{Bert for joint intent classification and slot
  filling}.
\newblock \bibinfo{journal}{\emph{arXiv preprint arXiv:1902.10909}}
  (\bibinfo{year}{2019}).
\newblock


\bibitem[Coucke et~al\mbox{.}(2018)]%
        {coucke2018snips}
\bibfield{author}{\bibinfo{person}{Alice Coucke}, \bibinfo{person}{Alaa Saade},
  \bibinfo{person}{Adrien Ball}, \bibinfo{person}{Th{\'e}odore Bluche},
  \bibinfo{person}{Alexandre Caulier}, \bibinfo{person}{David Leroy},
  \bibinfo{person}{Cl{\'e}ment Doumouro}, \bibinfo{person}{Thibault
  Gisselbrecht}, \bibinfo{person}{Francesco Caltagirone},
  \bibinfo{person}{Thibaut Lavril}, {et~al\mbox{.}}}
  \bibinfo{year}{2018}\natexlab{}.
\newblock \showarticletitle{Snips voice platform: an embedded spoken language
  understanding system for private-by-design voice interfaces}.
\newblock \bibinfo{journal}{\emph{arXiv preprint arXiv:1805.10190}}
  (\bibinfo{year}{2018}).
\newblock


\bibitem[Danilevsky et~al\mbox{.}(2020)]%
        {danilevsky2020survey}
\bibfield{author}{\bibinfo{person}{Marina Danilevsky}, \bibinfo{person}{Kun
  Qian}, \bibinfo{person}{Ranit Aharonov}, \bibinfo{person}{Yannis Katsis},
  \bibinfo{person}{Ban Kawas}, {and} \bibinfo{person}{Prithviraj Sen}.}
  \bibinfo{year}{2020}\natexlab{}.
\newblock \showarticletitle{A Survey of the State of Explainable AI for Natural
  Language Processing}. In \bibinfo{booktitle}{\emph{Proceedings of the 1st
  Conference of the Asia-Pacific Chapter of the Association for Computational
  Linguistics and the 10th International Joint Conference on Natural Language
  Processing}}. \bibinfo{pages}{447--459}.
\newblock


\bibitem[Dao et~al\mbox{.}(2021)]%
        {dao2021intent}
\bibfield{author}{\bibinfo{person}{Mai~Hoang Dao}, \bibinfo{person}{Thinh~Hung
  Truong}, {and} \bibinfo{person}{Dat~Quoc Nguyen}.}
  \bibinfo{year}{2021}\natexlab{}.
\newblock \showarticletitle{Intent detection and slot filling for Vietnamese}.
\newblock \bibinfo{journal}{\emph{arXiv preprint arXiv:2104.02021}}
  (\bibinfo{year}{2021}).
\newblock


\bibitem[Du et~al\mbox{.}(2019)]%
        {du2019techniques}
\bibfield{author}{\bibinfo{person}{Mengnan Du}, \bibinfo{person}{Ninghao Liu},
  {and} \bibinfo{person}{Xia Hu}.} \bibinfo{year}{2019}\natexlab{}.
\newblock \showarticletitle{Techniques for interpretable machine learning}.
\newblock \bibinfo{journal}{\emph{Commun. ACM}} \bibinfo{volume}{63},
  \bibinfo{number}{1} (\bibinfo{year}{2019}), \bibinfo{pages}{68--77}.
\newblock


\bibitem[Goo et~al\mbox{.}(2018)]%
        {goo2018slot}
\bibfield{author}{\bibinfo{person}{Chih-Wen Goo}, \bibinfo{person}{Guang Gao},
  \bibinfo{person}{Yun-Kai Hsu}, \bibinfo{person}{Chih-Li Huo},
  \bibinfo{person}{Tsung-Chieh Chen}, \bibinfo{person}{Keng-Wei Hsu}, {and}
  \bibinfo{person}{Yun-Nung Chen}.} \bibinfo{year}{2018}\natexlab{}.
\newblock \showarticletitle{Slot-gated modeling for joint slot filling and
  intent prediction}. In \bibinfo{booktitle}{\emph{Proceedings of the 2018
  Conference of the North American Chapter of the Association for Computational
  Linguistics: Human Language Technologies, Volume 2 (Short Papers)}}.
  \bibinfo{pages}{753--757}.
\newblock


\bibitem[Gunaratna et~al\mbox{.}(2021)]%
        {gunaratna2021using}
\bibfield{author}{\bibinfo{person}{Kalpa Gunaratna}, \bibinfo{person}{Vijay
  Srinivasan}, \bibinfo{person}{Sandeep Nama}, {and} \bibinfo{person}{Hongxia
  Jin}.} \bibinfo{year}{2021}\natexlab{}.
\newblock \showarticletitle{Using Neighborhood Context to Improve Information
  Extraction from Visual Documents Captured on Mobile Phones}. In
  \bibinfo{booktitle}{\emph{Proceedings of the 30th ACM International
  Conference on Information \& Knowledge Management}}.
  \bibinfo{pages}{3038--3042}.
\newblock


\bibitem[Gunaratna et~al\mbox{.}(2022)]%
        {gunaratna2022explainable}
\bibfield{author}{\bibinfo{person}{Kalpa Gunaratna}, \bibinfo{person}{Vijay
  Srinivasan}, \bibinfo{person}{Akhila Yerukola}, {and}
  \bibinfo{person}{Hongxia Jin}.} \bibinfo{year}{2022}\natexlab{}.
\newblock \showarticletitle{Explainable Slot Type Attentions to Improve Joint
  Intent Detection and Slot Filling}. In \bibinfo{booktitle}{\emph{Findings of
  the Association for Computational Linguistics: EMNLP 2022}}.
  \bibinfo{publisher}{Association for Computational Linguistics},
  \bibinfo{address}{Abu Dhabi, United Arab Emirates},
  \bibinfo{pages}{3367--3378}.
\newblock
\urldef\tempurl%
\url{https://doi.org/10.18653/v1/2022.findings-emnlp.245}
\showDOI{\tempurl}


\bibitem[Hakkani-T{\"u}r et~al\mbox{.}(2016)]%
        {hakkani2016multi}
\bibfield{author}{\bibinfo{person}{Dilek Hakkani-T{\"u}r},
  \bibinfo{person}{G{\"o}khan T{\"u}r}, \bibinfo{person}{Asli Celikyilmaz},
  \bibinfo{person}{Yun-Nung Chen}, \bibinfo{person}{Jianfeng Gao},
  \bibinfo{person}{Li Deng}, {and} \bibinfo{person}{Ye-Yi Wang}.}
  \bibinfo{year}{2016}\natexlab{}.
\newblock \showarticletitle{Multi-domain joint semantic frame parsing using
  bi-directional rnn-lstm.}. In \bibinfo{booktitle}{\emph{Interspeech}}.
  \bibinfo{pages}{715--719}.
\newblock


\bibitem[He et~al\mbox{.}(2020)]%
        {he2020syntactic}
\bibfield{author}{\bibinfo{person}{Keqing He}, \bibinfo{person}{Shuyu Lei},
  \bibinfo{person}{Yushu Yang}, \bibinfo{person}{Huixing Jiang}, {and}
  \bibinfo{person}{Zhongyuan Wang}.} \bibinfo{year}{2020}\natexlab{}.
\newblock \showarticletitle{Syntactic graph convolutional network for spoken
  language understanding}. In \bibinfo{booktitle}{\emph{Proceedings of the 28th
  International Conference on Computational Linguistics}}.
  \bibinfo{pages}{2728--2738}.
\newblock


\bibitem[Hemphill et~al\mbox{.}(1990)]%
        {hemphill1990atis}
\bibfield{author}{\bibinfo{person}{Charles~T Hemphill}, \bibinfo{person}{John~J
  Godfrey}, {and} \bibinfo{person}{George~R Doddington}.}
  \bibinfo{year}{1990}\natexlab{}.
\newblock \showarticletitle{The ATIS spoken language systems pilot corpus}. In
  \bibinfo{booktitle}{\emph{Speech and Natural Language: Proceedings of a
  Workshop Held at Hidden Valley, Pennsylvania, June 24-27, 1990}}.
\newblock


\bibitem[Lee et~al\mbox{.}(2018)]%
        {lee2018coupled}
\bibfield{author}{\bibinfo{person}{Jihwan Lee}, \bibinfo{person}{Dongchan Kim},
  \bibinfo{person}{Ruhi Sarikaya}, {and} \bibinfo{person}{Young-Bum Kim}.}
  \bibinfo{year}{2018}\natexlab{}.
\newblock \showarticletitle{Coupled representation learning for domains,
  intents and slots in spoken language understanding}. In
  \bibinfo{booktitle}{\emph{2018 IEEE Spoken Language Technology Workshop
  (SLT)}}. IEEE, \bibinfo{pages}{714--719}.
\newblock


\bibitem[Liu and Lane(2016)]%
        {liu2016attention}
\bibfield{author}{\bibinfo{person}{Bing Liu} {and} \bibinfo{person}{Ian Lane}.}
  \bibinfo{year}{2016}\natexlab{}.
\newblock \showarticletitle{Attention-Based Recurrent Neural Network Models for
  Joint Intent Detection and Slot Filling}.
\newblock \bibinfo{journal}{\emph{Interspeech 2016}} (\bibinfo{year}{2016}),
  \bibinfo{pages}{685--689}.
\newblock


\bibitem[Qin et~al\mbox{.}(2019)]%
        {qin2019stack}
\bibfield{author}{\bibinfo{person}{Libo Qin}, \bibinfo{person}{Wanxiang Che},
  \bibinfo{person}{Yangming Li}, \bibinfo{person}{Haoyang Wen}, {and}
  \bibinfo{person}{Ting Liu}.} \bibinfo{year}{2019}\natexlab{}.
\newblock \showarticletitle{A Stack-Propagation Framework with Token-Level
  Intent Detection for Spoken Language Understanding}. In
  \bibinfo{booktitle}{\emph{Proceedings of the 2019 Conference on Empirical
  Methods in Natural Language Processing and the 9th International Joint
  Conference on Natural Language Processing (EMNLP-IJCNLP)}}.
  \bibinfo{pages}{2078--2087}.
\newblock


\bibitem[Qin et~al\mbox{.}(2021)]%
        {qin2021co}
\bibfield{author}{\bibinfo{person}{Libo Qin}, \bibinfo{person}{Tailu Liu},
  \bibinfo{person}{Wanxiang Che}, \bibinfo{person}{Bingbing Kang},
  \bibinfo{person}{Sendong Zhao}, {and} \bibinfo{person}{Ting Liu}.}
  \bibinfo{year}{2021}\natexlab{}.
\newblock \showarticletitle{A co-interactive transformer for joint slot filling
  and intent detection}. In \bibinfo{booktitle}{\emph{ICASSP 2021-2021 IEEE
  International Conference on Acoustics, Speech and Signal Processing
  (ICASSP)}}. IEEE, \bibinfo{pages}{8193--8197}.
\newblock


\bibitem[Rudin(2019)]%
        {rudin2019stop}
\bibfield{author}{\bibinfo{person}{Cynthia Rudin}.}
  \bibinfo{year}{2019}\natexlab{}.
\newblock \showarticletitle{Stop explaining black box machine learning models
  for high stakes decisions and use interpretable models instead}.
\newblock \bibinfo{journal}{\emph{Nature Machine Intelligence}}
  \bibinfo{volume}{1}, \bibinfo{number}{5} (\bibinfo{year}{2019}),
  \bibinfo{pages}{206--215}.
\newblock


\bibitem[Socher et~al\mbox{.}(2013)]%
        {sst2}
\bibfield{author}{\bibinfo{person}{Richard Socher}, \bibinfo{person}{Alex
  Perelygin}, \bibinfo{person}{Jean Wu}, \bibinfo{person}{Jason Chuang},
  \bibinfo{person}{Christopher~D. Manning}, \bibinfo{person}{Andrew Ng}, {and}
  \bibinfo{person}{Christopher Potts}.} \bibinfo{year}{2013}\natexlab{}.
\newblock \showarticletitle{Recursive Deep Models for Semantic Compositionality
  Over a Sentiment Treebank}. In \bibinfo{booktitle}{\emph{Proceedings of the
  2013 Conference on Empirical Methods in Natural Language Processing}}.
  \bibinfo{publisher}{Association for Computational Linguistics},
  \bibinfo{address}{Seattle, Washington, USA}, \bibinfo{pages}{1631--1642}.
\newblock
\urldef\tempurl%
\url{https://www.aclweb.org/anthology/D13-1170}
\showURL{%
\tempurl}


\bibitem[Tjong Kim~Sang and De~Meulder(2003)]%
        {conll2003}
\bibfield{author}{\bibinfo{person}{Erik~F. Tjong Kim~Sang} {and}
  \bibinfo{person}{Fien De~Meulder}.} \bibinfo{year}{2003}\natexlab{}.
\newblock \showarticletitle{Introduction to the {C}o{NLL}-2003 Shared Task:
  Language-Independent Named Entity Recognition}. In
  \bibinfo{booktitle}{\emph{Proceedings of the Seventh Conference on Natural
  Language Learning at {HLT}-{NAACL} 2003}}. \bibinfo{pages}{142--147}.
\newblock
\urldef\tempurl%
\url{https://www.aclweb.org/anthology/W03-0419}
\showURL{%
\tempurl}


\bibitem[Wiegreffe and Pinter(2019)]%
        {wiegreffe2019attention}
\bibfield{author}{\bibinfo{person}{Sarah Wiegreffe} {and}
  \bibinfo{person}{Yuval Pinter}.} \bibinfo{year}{2019}\natexlab{}.
\newblock \showarticletitle{Attention is not not Explanation}. In
  \bibinfo{booktitle}{\emph{Proceedings of the 2019 Conference on Empirical
  Methods in Natural Language Processing and the 9th International Joint
  Conference on Natural Language Processing (EMNLP-IJCNLP)}}.
  \bibinfo{pages}{11--20}.
\newblock


\bibitem[Xu and Sarikaya(2013)]%
        {xu2013convolutional}
\bibfield{author}{\bibinfo{person}{Puyang Xu} {and} \bibinfo{person}{Ruhi
  Sarikaya}.} \bibinfo{year}{2013}\natexlab{}.
\newblock \showarticletitle{Convolutional neural network based triangular crf
  for joint intent detection and slot filling}. In
  \bibinfo{booktitle}{\emph{2013 ieee workshop on automatic speech recognition
  and understanding}}. IEEE, \bibinfo{pages}{78--83}.
\newblock


\bibitem[Zhang et~al\mbox{.}(2019)]%
        {zhang2019joint}
\bibfield{author}{\bibinfo{person}{Chenwei Zhang}, \bibinfo{person}{Yaliang
  Li}, \bibinfo{person}{Nan Du}, \bibinfo{person}{Wei Fan}, {and}
  \bibinfo{person}{S~Yu Philip}.} \bibinfo{year}{2019}\natexlab{}.
\newblock \showarticletitle{Joint Slot Filling and Intent Detection via Capsule
  Neural Networks}. In \bibinfo{booktitle}{\emph{Proceedings of the 57th Annual
  Meeting of the Association for Computational Linguistics}}.
  \bibinfo{pages}{5259--5267}.
\newblock


\bibitem[Zhang et~al\mbox{.}(2020)]%
        {zhang2020graph}
\bibfield{author}{\bibinfo{person}{Linhao Zhang}, \bibinfo{person}{Dehong Ma},
  \bibinfo{person}{Xiaodong Zhang}, \bibinfo{person}{Xiaohui Yan}, {and}
  \bibinfo{person}{Houfeng Wang}.} \bibinfo{year}{2020}\natexlab{}.
\newblock \showarticletitle{Graph lstm with context-gated mechanism for spoken
  language understanding}. In \bibinfo{booktitle}{\emph{Proceedings of the AAAI
  Conference on Artificial Intelligence}}, Vol.~\bibinfo{volume}{34}.
  \bibinfo{pages}{9539--9546}.
\newblock


\end{thebibliography}

\end{document}